\documentclass[10pt,twocolumn,letterpaper]{article}

\usepackage{cvpr}              

\usepackage[accsupp]{axessibility}  
\usepackage{graphicx}
\usepackage{amsmath}
\usepackage{amssymb}
\usepackage{booktabs}
\usepackage{xcolor}

\usepackage{lipsum}  
\usepackage{hhline}
\usepackage{caption}
\usepackage{subcaption}
\usepackage{mathrsfs}

\usepackage[pagebackref,breaklinks,colorlinks]{hyperref}

\usepackage[capitalize]{cleveref}
\crefname{section}{Sec.}{Secs.}
\Crefname{section}{Section}{Sections}
\Crefname{table}{Table}{Tables}
\crefname{table}{Tab.}{Tabs.}

\def\cvprPaperID{81} 
\def\confName{CVPR}
\def\confYear{2022}

\begin{document}

\title{DRHDR: A Dual  branch Residual Network for Multi-Bracket High Dynamic Range Imaging}

\author{Juan Mar\'in-Vega$^{1,2}$ \and Michael Sloth$^{2}$ \and Peter Schneider-Kamp$^{1}$ \and Richard Röttger$^{1}$\\
\and
$^{1}$~Department of Mathematics and Computer Science (IMADA), University of Southern Denmark\\
$^{2}$~Esoft Systems\\
{\tt\small msl@esoft.com \{marin,petersk,roettger\}@imada.sdu.dk }}
\maketitle

\begin{abstract}
	We introduce DRHDR, a Dual branch Residual Convolutional Neural Network for Multi-Bracket HDR Imaging. To address the challenges of fusing multiple brackets from dynamic scenes, we propose an efficient dual branch network that operates on two different resolutions. The full resolution branch uses a Deformable Convolutional Block to align features and retain high-frequency details. A low resolution branch with a Spatial Attention Block aims to attend wanted areas from the non-reference brackets, and suppress displaced features that could incur on ghosting artifacts. By using a dual branch approach we are able to achieve high quality results while constraining the computational resources required to estimate the HDR results.
\end{abstract}
{\let\thefootnote\relax\footnotetext{%
\hspace{-5mm}
$^{3}$~Project available at \url{https://github.com/drhdr-user/drhdr}
}}


\section{Introduction}
\label{sec:intro}
Digital camera sensors have limited capabilities at capturing the rich ranges of luminance values of natural images. In order to produce images that resemble what the human eye can see, common HDR methods utilize multiple frames with different exposure values, to produce a final HDR image that has a higher fidelity with respect to the original scene. Nevertheless, HDR solutions still have to deal with the challenges that arise when merging brackets under different light conditions: Saturated regions and noise. Moreover, camera motion and moving objects can also affect the final estimated image by introducing ghosting artifacts. 

Several solutions have been proposed over time to solve these challenges: pixel rejection approaches\cite{khanGhostRemovalHigh2006,grosch2006fast,gallo2009, minHistogramBasedGhost2009,pece2010bitmap,leeGhostFreeHighDynamic2014,ohRobustHighDynamic2015} that select and minimize the contributions of areas that contain motion objects or misalignment, but also pixel registration techniques\cite{bogoniExtendingDynamicRange2000,kang2003,jinnoMotionBlurFree2008,zimmerFreehandHDRImaging2011,senRobustPatchbasedHdr2012} for selecting the areas that would provide the best content for the final solution.

With the recent increase on computational capabilities and the improved collection of data, several deep learning systems trained in a supervised fashion have improved HDR estimation results with a big margin with respect of previous approaches. These deep learning techniques are able to work on a feature space\cite{kalantariDeepHighDynamic2017,wuDeepHighDynamic2018}, reject and align regions based on reference frames\cite{yan_attention_guided_2019, wuDeepHighDynamic2018} and on a non-determined number of frames\cite{catley2022} for producing high fidelity output estimations.

\begin{figure}[!t]\centering
	\includegraphics[width=.49\textwidth]{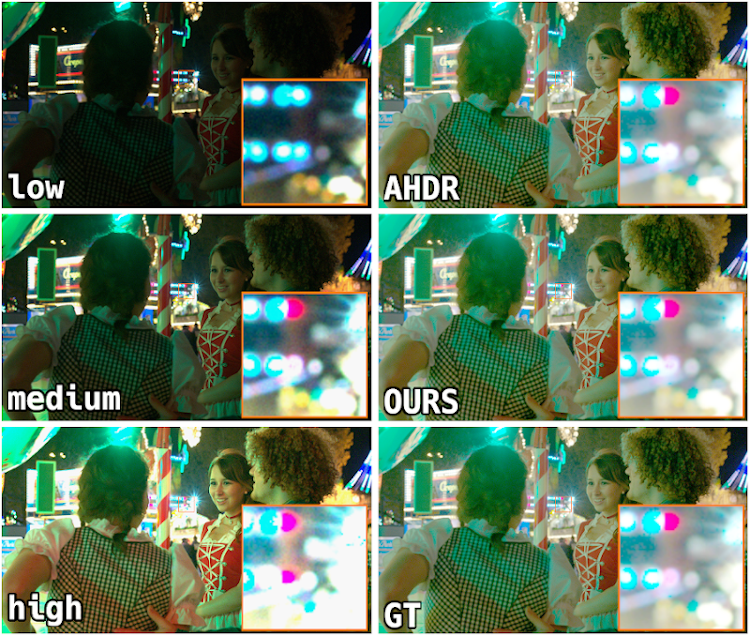}
	\caption{\textbf{Qualitative comparison} from our validation split comparing the AHDR Baseline with our technique. Our solution effectively merges the three input brackets and provide high quality outputs, free of noise or saturated areas. Moreover, effectively achieves higher fidelity on the zoomed areas when compared with the ground truth sample.}
	\label{fig:qualitative-teaser}
\end{figure}

However, despite the impressive quality improvements, these techniques still are far from optimal in terms of their computational requirements. On this basis, the NTIRE 2022 HDR Challenge\cite{perezpellitero22} aims to stimulate research on HDR imaging techniques, with an emphasis on efficient solutions. Two tracks are proposed in the competition. Track 1: Fidelity (Low-complexity constrain), and Track 2: Low-Complexity (Fidelity constrain). In Track 1, participants are asked to optimize fidelity scores (PSNR and  PSNR-$\mu$) while keeping complexity under 200 GMACs. In Track 2, participants are asked to minimize the complexity of their solutions (GMACs and runtime) while achieving at least the same fidelity scores (PSNR and  PSNR-$\mu$) as the baseline method AHDR\cite{yan_attention_guided_2019}. In this article, we describe our proposed solution for Track 2.

Inspired by previous works \cite{GharbiCBHD17,ShahamG0SM21,Moran20,wuDeepHighDynamic2018,wangDeepHighResolutionRepresentation2020} that combine multiple-resolution transformations, either with U-Nets, low resolution branches with operations that are locally  smooth when translated to full resolution results; our network aims to leverage on a spatially reduced feature space for alleviating the number of computations that would require operating only on full resolution. Hence, our network is built with two branches that operate on different resolutions. Branch $b_0$ operates at full resolution while $b_1$ operates at a fourth of the original resolution. We also incorporate previously successful ideas as the Spatial Attention, Deformable Convolutions, and Dilated Residual Dense Blocks. This combination provides a faster, and less computationally expensive framework, with an increase in quality compared to baseline solutions. We summarize our contributions as follows:
\begin{itemize}
	\setlength{\itemsep}{0pt}
	\item We propose a dual branch system that works at two different resolution for reducing the computational complexity while still improving on quality metrics.
	
	\item We asses the benefits of the components used on each of the branches, a careful explanation of the different training options, followed by quantitative and qualitative results.
	
\end{itemize}

\begin{figure*}[!t]\centering
	\includegraphics[width=.9\textwidth]{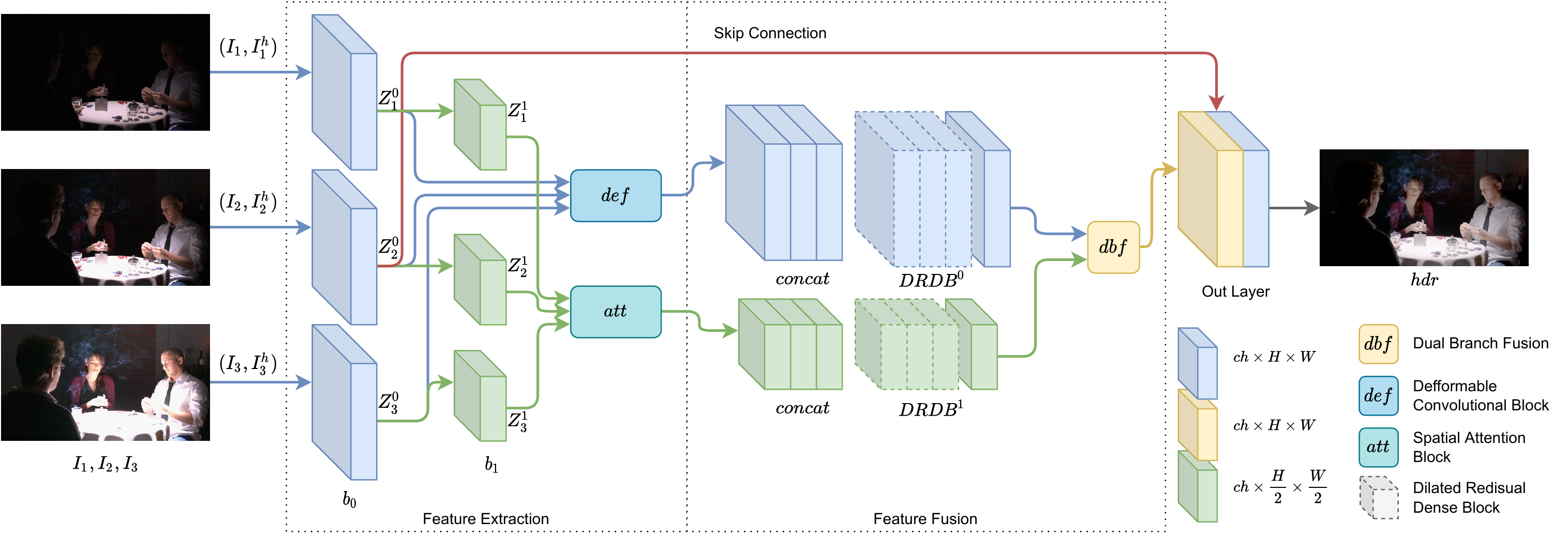}
	\caption{\textbf{Diagram of our method}. Two main branches. $b_0$ works at full resolution and uses a  Deformable Convolutional Block while $b_1$ works at fourth of the original resolution and uses a spatial attention module. Both branches are supplemented with Dilated Residual Dense Blocks and a Dual Branch Fusion Block. }
	\label{fig:diagram}
\end{figure*}
\section{Related work}
\label{sec:related}
For High Dynamic Range Imaging with dynamic scene we have a series of major challenges to solve. Restoring highly saturated regions, noise, misalignment from camera motion on the LDR brackets and dealing with entities in movement. The latest being the most challenging one since foreground objects might occlude regions of interest, but also creating ghosting artifacts. These ghosting artifacts have a similar nature to the ones created by non-aligned brackets, but they can be manifested in a much more severe manner, since objects can have big displacements across the different brackets.

Several approaches have been proposed to tackle these challenging scenarios. Historically, these approaches were based on either rejection or registration. Pixel rejection approaches aim to detect and minimize the contribution of regions that contain moving objects in order to reduce the probability of ghost areas on the final merged output. Pixels can be rejected after an iterative process of computing weights per pixel and the probability that a pixel is capturing the background\cite{khanGhostRemovalHigh2006}, predicting and measuring color difference from inputs\cite{grosch2006fast,gallo2009}, by multi-level thresholding of histograms\cite{minHistogramBasedGhost2009} or by detecting clusters of moving pixels by using binary operations\cite{pece2010bitmap}. Lately Lee~\etal~\cite{leeGhostFreeHighDynamic2014} and Oh~\etal~\cite{ohRobustHighDynamic2015} proposed using rank minimization for aligning LDR inputs and ghost region detection. 
Pixel or Patch registration methods aim to estimate HDR outputs by localizing the best regions on the LDR inputs. Bogoni\cite{bogoniExtendingDynamicRange2000} uses Laplacian Pyramids for salience detection, selection and fusion.  Kang~\etal~\cite{kang2003} selects the best pair of LDR based on pixel brightness distribution, and uses hierarchical homography for registration and compensation of pixel motion. The final output incorporates portions from the inputs that are not saturated, and that contain the best details. Jinno and Okuda\cite{jinnoMotionBlurFree2008} estimate displacement, occlusion and saturation based on irradiance values of each pixel with a Markov random field model, to produce blur-free results. Zimmer~\etal~\cite{zimmerFreehandHDRImaging2011} align the input LDR with an energy-based optic flow method that takes into account the varying exposure conditions to create displacement fields with subpixel precition. Sen~\etal~\cite{senRobustPatchbasedHdr2012} optimized jointly alignment and reconstruction using a patch-based energy-minimization formulation.

Deep learning approaches have significantly improve results compared to previous techniques. To do so, supervised learning approaches require datasets with input output pairs. Kalantari and Ramamoorthi~\cite{kalantariDeepHighDynamic2017} introduced a convolutional neural network for alignment and merging and collected a dataset of LDR inputs and an HDR output. To enforce the misalignment of the inputs, they replace the original low and high exposure LDR inputs with a pair of low, high LDR inputs that are not aligned with the reference middle bracket. Wu~\etal~\cite{wuDeepHighDynamic2018} explores the utilization of U-Nets\cite{RonnebergerFB15} for better exploiting deep representations and perform alignment on the feature space. Yan~\etal~\cite{yan_attention_guided_2019} introduced the use of Spatial Attention Layers, that based on a reference bracket, can detect and suppress regions of the non-reference brackets to produce ghost-free estimations. For NTIRE 2021 Challenge\cite{NTIRE2021Challenge2021}, Liu~\etal~\cite{liu_adnet_2021} proposed the use of Deformable Convolutions for better alignment of brackets. Previously, the use of Deformable Convolutions had already shown promising improvements on discriminative methods\cite{zhuDeformableConvNets2017,zhuDeformableConvNetsV22021} as well as for video frame alignment\cite{wang_edvr_2019}. Recently, Yan~\etal~\cite{yanDualAttentionGuidedNetworkGhostFree2022} improves upon their previous contributions, by providing a Dual Attention mechanism and a channel Attention Mechanism. Finally, Catley-Chandar~\etal~\cite{catley2022} introduces a system that jointly aligns and merges frames based on uncertainty-drive attention maps, and a progressive multi-stage image fusion that can work with an arbitrary number of input brackets. Other deep learning based works have explored unsupervised HDR fusion\cite{prabhakarDeepFuseDeepUnsupervised2017}, reinforcement learning for bracket selection\cite{wangLearningReinforcedAgent2020} and the use of adversarial training for better hallucination of missing content\cite{niuHDRGANHDRImage2021}.


\begin{figure}[!t]\centering
	\includegraphics[width=.49\textwidth]{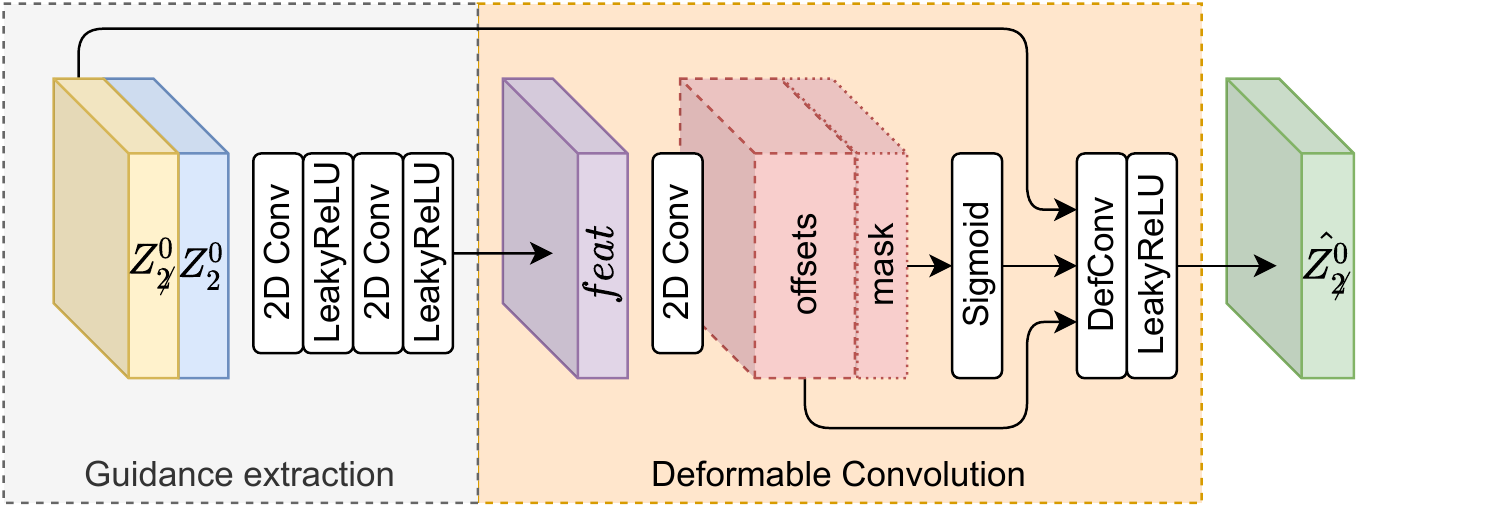}
	\caption{\textbf{Deformable Convolutional Block}: A stack of 2D Convolutions followed by a LeakyReLU are in charge of extracting the guidance features from $[Z^0_{\not2},Z^{0}_2]$. These features are used for generating the offsets and the mask for the deformable convolution. The resultant non-reference features $\hat{Z^0_{\not2}}$ are obtained applying a Deformable Convolution over $Z^0_{\not2}$ together with the mask and the given offsets.}
	\label{fig:deformable}
\end{figure}

\section{Methodology}
\label{sec:methodology}

\subsection{Architecture}
\label{sec:methodology}
The overall architecture comprises two routes, one branch that operates at full resolution, and a second branch that operates at a  fourth of the original resolution. We define a fixed number of channels for both branches $ch=42$. The full resolution branch adopts a Deformable Convolutional Block~\cite{liu_adnet_2021} while the low resolution branch adopts Spatial Attention~\cite{yan_attention_guided_2019}. Since both LDR and gamma adjusted contribute to detecting and misalignments \cite{yan_attention_guided_2019,wuDeepHighDynamic2018,liu_adnet_2021}, we opt for using LDR and gamma adjusted as inputs for both branches. Given 3 input LDR brackets $I_i, i = 1,2,3$ the network input is composed as $[I_i, I^{h}_i], i = 1,2,3$, a channel-wise concatenation of the original $I$ and its gamma corrected $I^h$. Where $I_2$ is the reference middle bracket, and  $I_i, i = 1,3$ the non-reference ones.

\begin{figure}[!t]\centering
	\includegraphics[width=.37\textwidth]{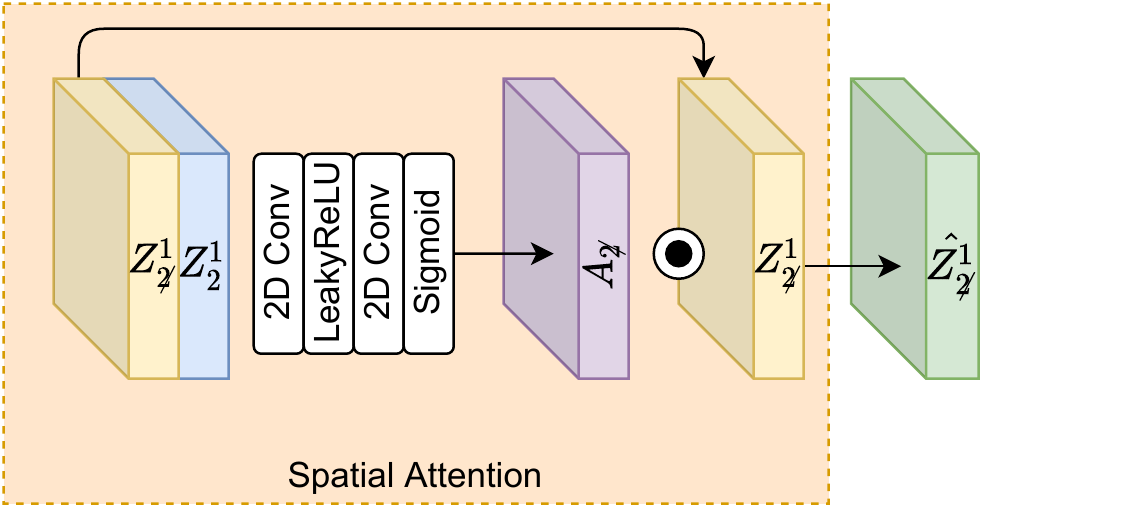}
	\caption{\textbf{Spatial Attention Block}: Attention maps $A_{i}, i=1,3$ are obtained from $[Z^1_{\not2},Z^{1}_2]$ after 2 2D Convolutions follows by a LeakyReLU and a Sigmoid activation respectively. $\hat{Z^1_{\not2}}$ is finally obtained by a point-wise multiplication on $Z^1_{\not2}$.}
	\label{fig:spatial}
\end{figure}

\subsubsection{Full Resolution Branch}
The full resolution branch starts with an input layer $b_0$ composed by a $3\times3\times42$ 2D Convolution and a Leaky ReLU activation. This layer is responsible of encoding the 6 channels input $[I_i, I^{h}_i]$ onto 42 channels features $Z^{0}_i$ for every bracket $i = 1,2,3$. The input layer $b_0$ is followed by a Guided Deformable Convolutional Block $def$ and a Dilated Residual Dense Block\cite{yan_attention_guided_2019,zhangResidualDenseNetwork2018a} $DRDB^{0}$.

\paragraph{Deformable Convolutional Block}
Similarly to ADNET~\cite{liu_adnet_2021}, the Deformable Convolutional Block $def$ is responsible for aligning images in the feature space. Deformable convolutions improve the ability to model geometric transformations. Given that their spatial support is higher than regular convolutions~\cite{zhuDeformableConvNetsV22021}, it can be effective when applied to high resolution, and not so deep, feature representations, accounting for high frequency details and finer alignments.  Our $def$ is a light-weight adaptation of the module used in ADNET\cite{liu_adnet_2021} and EDVR\cite{wang_edvr_2019}. Instead of a full Pyramid, Cascading and Deformable PCD convolution approach, that works at 3 different resolutions, ours work only on full resolution features. \Cref{fig:deformable} illustrates the Deformable Convolutional Block. A stack of 2D Convolutions followed by a LeakyReLU are in charge of extracting the guidance features from $[Z^0_{\not2},Z^{0}_2]$. These features are used for generating the offsets and the mask for the deformable convolution from which the final $\hat{Z^0_{\not2}}$ is obtained. In particular, 2 $def$ modules are defined, one for each set of non-reference features $l^{i}_0, i=1,3$.

\subsubsection{Low Resolution Branch}
The low resolution branch $b_1$ reduces the spatial dimensionality of $Z^{0}_i$ by half on each edge after applying a strided 2D Convolution and LeakyReLU to produce $Z^{1}_i$, with a shape of $B\times42\times \frac{H}{2}\times \frac{W}{2}$, where $B$ is the batch size, and $H,W$ the input resolution. Features $Z^{1}_i$ are then processed through a Spatial Attention Block and a Dilated Residual Dense Block\cite{yan_attention_guided_2019,zhangResidualDenseNetwork2018a} $DRDB^{1}$. 

\paragraph{Spatial Attention}
The Spatial Attention Block~\cite{yan_attention_guided_2019} $att$ allows the network to extract features of particular areas of the inputs. It suppresses activations from the features by performing point-wise multiplications. As depicted in~\Cref{fig:spatial}, the attention maps $A_{i}, i=1,3$ are obtained from $[Z^1_{\not2},Z^{1}_2]$ after 2 2D Convolutions follows by a LeakyReLU and a Sigmoid activation respectively. $\hat{Z^1_{\not2}}$ is finally obtained by

$$ \hat{Z^1_{i}} = A_{i} \circ Z^1_{i}, i=1,3  $$ 

Two $att$ blocks are defined, for each set of non-reference features $Z^1_{i}, i=1,3$

\subsubsection{Dual Branch Fusion}
The outputs of $DRDB^{0}$ and $DRDB^{1}$ are the features from the branches at different resolutions. $dbf$ is responsible for upscaling the set of low-resolution features from $b_1$ to match the resolution of $b_0$. After this operation, both sets of features are concatenated and fused through a 2D Convolution and a LeakyReLU.

\subsubsection{Output Layer}
The output layer, uses a global skip dense connection. In particular, the output from $dbf$ and full resolution encoded features $Z^{0}_2$ from the reference input $[I_2, I^{h}_2]$ and apply 2D Convolution and Leaky ReLU followed by another 2D Convolution and a final ReLU.

\begin{figure*}[!t]
	\begin{subfigure}{.49\textwidth}\centering
		\includegraphics[width=1\textwidth]{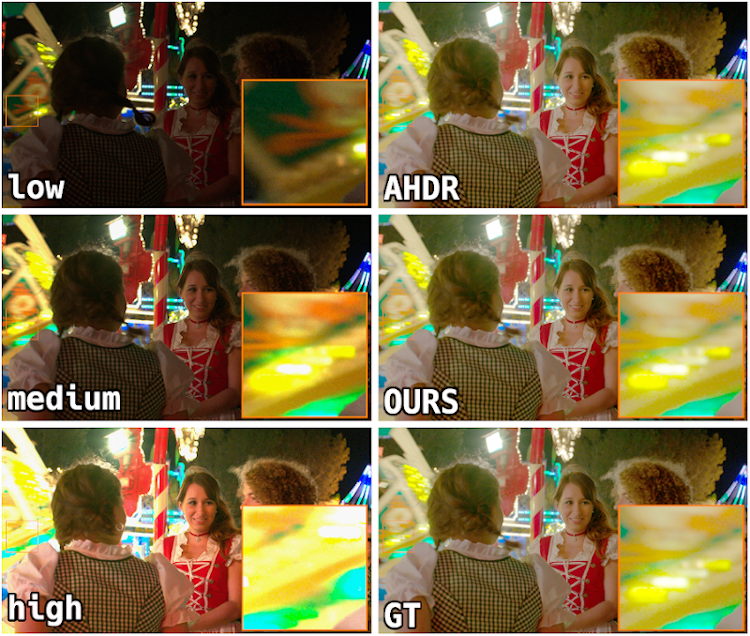}
		\caption{}
	\end{subfigure}
	\begin{subfigure}{.49\textwidth}\centering
		\includegraphics[width=1\textwidth]{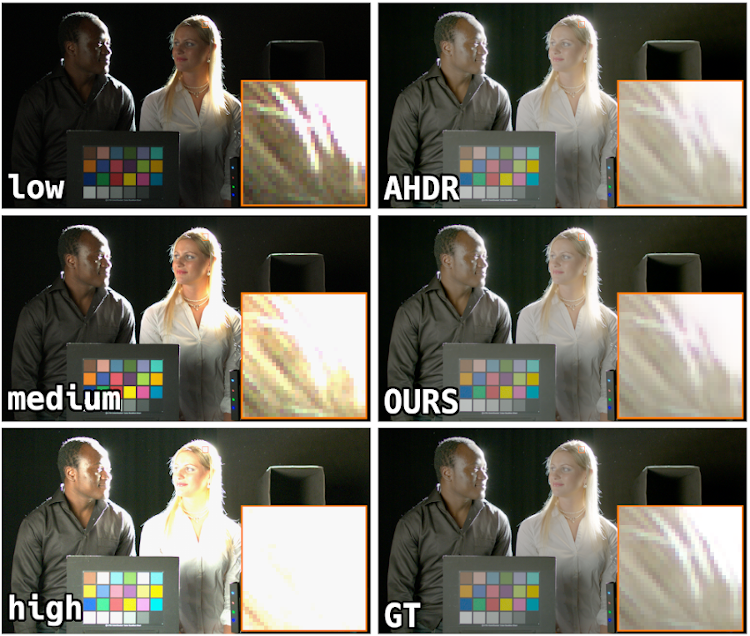}
		\caption{}
		\label{fig:qualitative-big-b}
	\end{subfigure}
	\caption{\textbf{Qualitative comparison} from our validation split comparing the AHDR Baseline with our technique. Our technique achieves higher accuracy on saturated areas but also produce more clear details.}
	\label{fig:qualitative-big}
\end{figure*}

\section{Experiments}
\label{sec:experiments}
In this section we elaborate on the most important details of the training process and the different experiments that lead to our final solution. We also cover comparisons against baselines. We refer to ``Baseline$_{in}$'' for the AHDR implementation trained and evaluated under the same regime as our variants and final solution. We refer to ``Baseline'' when comparing against the AHDR baseline results from the NTIRE 2022 HDR Challenge\cite{perezpellitero22}. 

\subsection{Training Details}
\subsubsection{Loss Function}
Given that training the network on the tonemapped images is more effective\cite{kalantariDeepHighDynamic2017,yan_attention_guided_2019,liu_adnet_2021}, we utilize $l_1$ loss over the  estimated $\hat{I}^{h}$ and ground truth $I^{GT}$ images. We normalize using using the $\mu$-law tonemapping\cite{liu_adnet_2021}:

\begin{equation}
 mu(x) = \frac{log(1 + \mu~x)}{log(1 + \mu)}.
\end{equation}
\label{eq:1}

We utilize $\mu=5000$ and prior to the tonemapping, we apply $tanh$ normalization using the 99 percentile from the estimated $\hat{I}^{h}$.  

\subsubsection{Data}
We split the dataset for Training and Validation. We randomly select 250 samples from the dataset for validation and use the rest for training. The images for training are cropped on $250\times250$ patches with stride of $250px$. 

\subsubsection{Evaluation Metrics}
Evaluation is measured on two metrics. Peask Signal to Noise Ratio PSNR that is computed directly on linear HDR estimations. The second metric is PSNR-$\mu$, which is computed over the estimated images after tonemapping with the $\mu$-law defined in \cref{eq:1}. The implementation of these metrics are provided by the NTIRE 2022 HDR Challenge\cite{perezpellitero22} administrators.

\subsubsection{Training Phases}
The network is trained using Adam optimizer with $\beta_1=0.9$ and $\beta_1=0.999$. The network is trained for 300 epochs in 3 different phases (Figure~\ref{fig:lr}). The training is conducted on a single RTX A6000 with 48GB using a batch size of 28 for approximately 5 days.\footnote{By using Nvidia Automatic Mixed Precision AMP during training, memory and time requirements can be reduced by $\sim$~half. However, we report results from the model submited to the challenge, that was trained without AMP. }

\paragraph{First training phase}
The network is initially trained for 140 epochs with $lr=0.0005$ and 10 epochs with $lr=0.0001$. For speeding up the training process, in this phase we use 50\% of the available training data, and randomly replace it at the start of each epoch.

\paragraph{Second training phase}
The network is trained for 50 epochs using all training data with an starting $lr=0.0001$  that decays following Equation \ref{eq:lrdecay}:

\begin{equation} \label{eq:lrdecay}
	decay(x) = x\frac{1 - n}{N}.
\end{equation}

where $x$, $n$ and $N$ are initial learning rate, the epoch and the total of epochs of the current phase.

\paragraph{Last training phase}
In this phase we train for the remaining 100 epochs using all training data with a decaying learning rate that starts from $lr=0.00005$.

\subsubsection{Data augmentation}
We randomly apply one of the following operations during training: vertical flip, horizontal flip, $90^\circ$ rotation, or nothing.

\subsubsection{Testing}
Testing is performed on a RTX Quadro 6000 with 24GB. Our method has a measured runtime of $\approx$ 0.75 seconds at processing images of 1060x1900 pixels. Our method has a peak memory consumption of $\approx$14394MiB using \texttt{float32} precision.
%


\subsubsection{Variants}
\label{section:variants}

\begin{table}[!t]
	\begin{center}
		\begin{tabular}{l|cccc}
			\textbf{Method}    			
			& \textbf{PSNR}$\uparrow$
			& \textbf{PSNR-$\mu$}$\uparrow$
			& \textbf{\#weights}$^\downarrow$
			& \textbf{GMACs}$\downarrow$\\ \hline
			Variant A				& 40.95 & 35.21 & 1345323 & 2146.68 			\\ \hline
			Variant B				& 40.83 & 35.17 & 1338087 & \textbf{1526.83} 	\\ \hline	
			Ours			 		& \textbf{41.03} & \textbf{35.22} & \textbf{1222035} & 1769.85 \\ \hhline{=|====}
			Baseline$_{in}$ 		& 40.61 & 35.14 & 1441283 & 2916.92				\\

		\end{tabular}
	\end{center}
	\caption{\textbf{Quantitative Results} of the different variants on our validation split. Baseline$_{in}$ refers to the Baseline implementation from AHDR\cite{yan_attention_guided_2019} trained and tested under the same regime as the rest. Please refer to \cref{section:variants} for more details on the different variants.}
	\label{table:local-results}
\end{table}

\paragraph{Variant A}
We define a Variant A based on the AHDR\cite{yan_attention_guided_2019} implementation where we incorporate a low resolution branch with a second Spatial Attention Block and the same number of DRDB blocks ($=6$) as in the full resolution branch. In order not to double the number of model weights, we reduce the number of channels on each convolution from $ch=64$ to $ch=36$. We expect the low resolution branch to compensate the reduction on the number of channels per convolution, but also to further reduce the number of GMACs by performing operations on a spatially reduced feature space.

\paragraph{Variant B}
Based on Variant A, we replace the Spatial Attention Block from the full resolution branch with a Deformable Convolutional Block. We incorporate the Deformable Convolutional Block  only on the full resolution branch, to perform a finer alignment and retain high frequency details. On the other hand, the Spatial Attention Block  on the low resolution branch, would be responsible for the suppression of non-wanted features from the non-reference brackets. We expect this suppression to still be  effective on a spatially reduced feature space. Moreover, since the Deformable Convolutional Block performs a higher number operations (and has a higher number of trainable parameters) compared to the Spatial Attention Block, we compensate it by reducing the number of Dilated Residual Dense Blocks on the full resolution branch, in this case $=3$ DRDB on the full resolution branch, $=6$ on the low resolution branch.

\begin{table}[!t]
	\begin{center}
		\begin{tabular}{l|cc}
			\textbf{Method}    			
			& \textbf{PSNR}$\uparrow$
			& \textbf{PSNR-$\mu$}$\uparrow$
			\\ \hline
			
			Variant A				& 38.38 & 36.89\\ \hline 
			Variant B				& 38.35 & 36.87\\ \hline
			Ours			 		& \textbf{38.50} & \textbf{36.91} \\ 
		\end{tabular}
	\end{center}
	\caption{\textbf{Quantitative Results} of the different variants on the validation split from NTIRE 2022 HDR Challenge\cite{perezpellitero22}.}
	\label{table:validation-results}
\end{table}

\begin{table}[!t]
	\begin{center}
		\begin{tabular}{l|cc}
			\textbf{Loss}    			
			& \textbf{PSNR}$\uparrow$
			& \textbf{PSNR-$\mu$}$\uparrow$\\ \hline			
			L1 	& \textbf{43.85} & 35.07 \\ \hline
			L1$_{tanh}$ 	& 40.61 & \textbf{35.14} \\ 			
		\end{tabular}
	\end{center}
	\caption{\textbf{Quantitative Results} of the Base L1 Loss vs L1$_{tanh}$ Loss on the Baseline$_{in}$ model. L1$_{tanh}$ provides better PSNR-$\mu$ results. Results computed on our validation split.}
	\label{table:results-loss}
\end{table}

\begin{table}[!t]
	\begin{center}
		\begin{tabular}{l|cc}
			\textbf{Method}    			
			& \textbf{PSNR}$\uparrow$
			& \textbf{PSNR-$\mu$}$\uparrow$\\ \hline
			Variant B				& 44.02 & 35.14
		\end{tabular}
	\end{center}
	\caption{\textbf{Quantitative Results} of the combining Base L1 Loss vs L1$_{tanh}$ Loss on the Variant B. Results computed on our validation split.}
	\label{table:local-results-combined-loss}
\end{table}

\paragraph{Final Architecture}
\Cref{table:local-results} reports results on our validation split. We also include results of Baseline$_{in}$, the Baseline implementation from AHDR\cite{yan_attention_guided_2019} trained and tested under the same regime as Variant A, B and the final architecture \textit{Ours}. It can be seen that Variant A and B, perform above Baseline$_{in}$. Also, Variant B has similar number of trainable parameters but performs $\approx25\%$ less operations than Variant A, and $\approx45\%$ less operations than Baseline$_{in}$. Still, Variant A achieves slightly better quality performance than Variant B. Motivated by these results, we adopt some changes to define the final architecture, based on Variant B we increase the number of channels from 36 to 42 and define an equal number of DRDB layers on both branches $=5$ (+2 on the full resolution branch, -1 on the low resolution branch), and reduce the number of expanding channels in each DRDB from 36 to 21. This leads to a model with a further reduced footprint, in terms of trainable weights, but a slightly more computationally expensive compared to Variant B. \cref{table:local-results} shows that this final architecture balances out to a smaller but a slightly better architecture overall. We report on \Cref{table:validation-results} results on the validation split from the challenge. Notice, these architectures perform above the Baseline implementation from NTIRE 2022 HDR Challenge\cite{perezpellitero22}. It can be seen that the final solution is still more powerful in PSNR and PSNR-$\mu$ than Variant A and B. We notice that with this validation split, PSNR-$\mu$ results are higher and PSNR results are lower, concluding that our internal validation split reports slightly optimist results for PSNR.

\begin{figure}[!t]\centering
	\includegraphics[width=.35\textwidth]{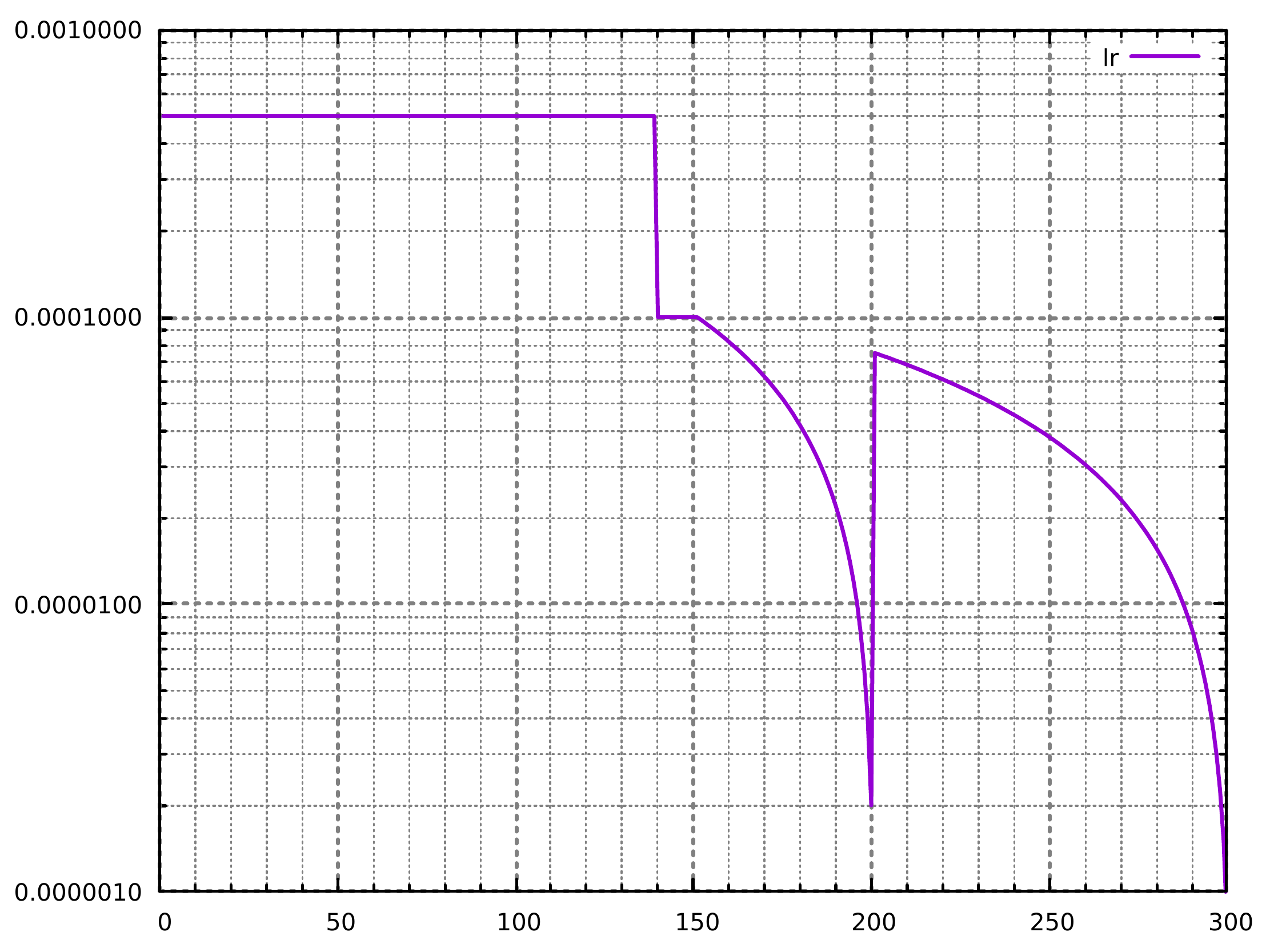}
	\caption{\textbf{Learning rate} training strategy. We perform training on 3 different phases. During the first phase we keep a static training rate of 0.0005 and change it to 0.0001 for the last 10 epochs. Then we train for 2 more phases utilizing a learning rate decay method from Equation \ref{eq:lrdecay}.}
	\label{fig:lr}
\end{figure}

\begin{figure*}[!t]
	\includegraphics[width=1\textwidth]{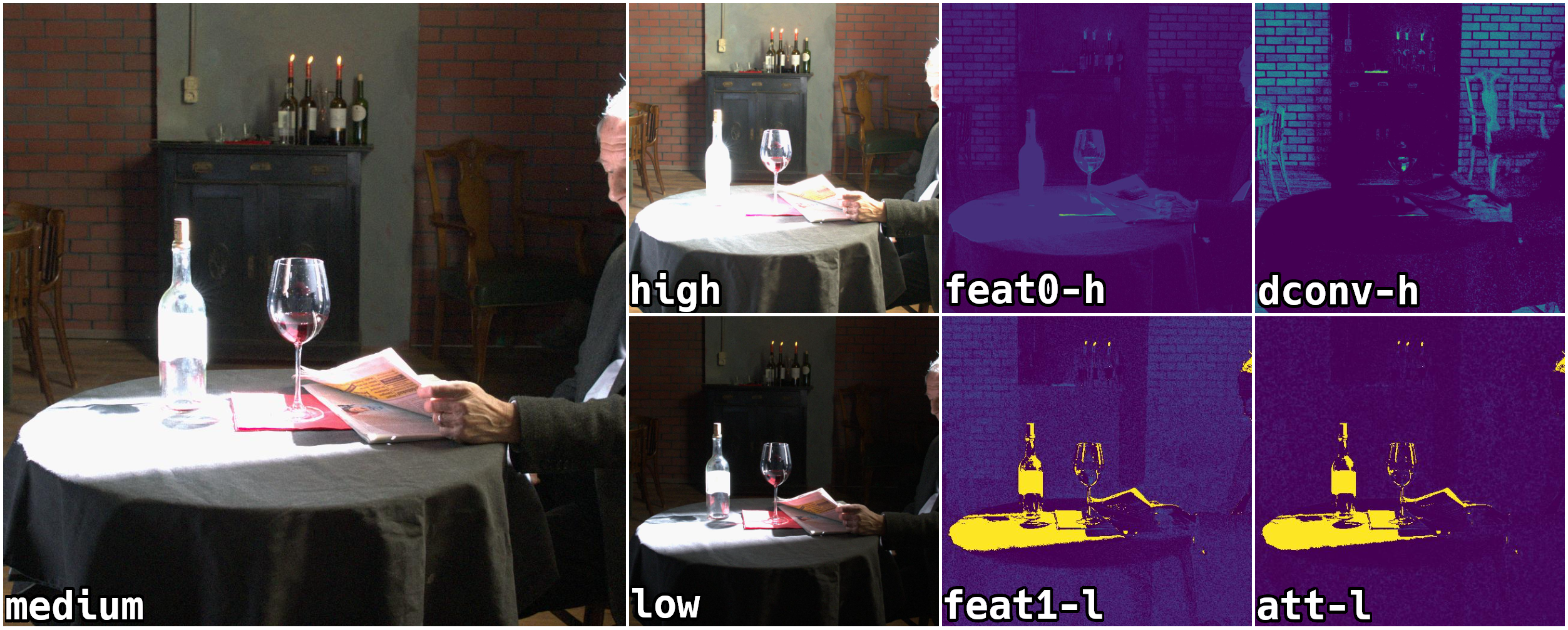}
	\caption{\textbf{Qualitative representation} of the behavior from the main components of our system. \texttt{feat0-h} is a single activation channel from $Z^{0}_3$, and \texttt{dconv-h} is the same channel result after applying the Deformable Convlutional Block. \texttt{Dconv-h} represents the effective minimization of details from the over-exposed regions of the table, while enhancing details from the background wall. At the bottom, \texttt{feat1-l} is a single activation channel from $Z^{0}_1$ while \texttt{att-l} is the same channel after applying the Spatial Attention Block. \texttt{Att-l} despite being  at a fourth of the original resolution, is able to retain the well-exposed details from the table, while  representing an accurate reduction of the over-exposed details from the background wall.}
	\label{fig:qualitative-maps}
\end{figure*}

\subsection{Loss Function}
In order to assess the performance of the loss function, we train a Baseline$_{in}$ implementation with L1 loss, and another using L1$_{tanh}$ where before the tonemapping, we apply $tanh$ normalization using the 99 percentile from the estimated $\hat{I}^{h}$. We report the results on \Cref{table:results-loss}. In this case we see an increase of $\approx$3dB in PSNR using the L1, but an increase of $\approx$.7dB on PSNR-$\mu$ by using the L1$_{tanh}$. After observing these results we proceed training Variant B with both losses. We report results on \Cref{table:local-results-combined-loss} on our validation set. We see an improvement on PSNR but a slight degradation on PSNR-$\mu$. We notice that PSNR-$\mu$ metric is probably more challenging to increase compared to the regular PSNR: Out of 53 groups contestants, only 14 achieve higher PSNR-$\mu$ than the reference implementation. Thus, we decide to proceed only with L1$_{tanh}$ instead of combining both.

\subsection{Learning Rate}
Finally, we perform a comparison based on the initial learning rate. We compare the performance of our final solution after 150 epochs, following the specifications of \cref{fig:lr} (Only first phase), against a model trained with a learning rate of 0.0001. \Cref{table:learning-rate} reports the results of this experiment. By training with a slightly higher learning rate our model achieves higher performance. On the other hand, by training with a lower learning rate, the model seems to stabilize on a sub-optimal local minima based on the fact that after epoch 125 the metrics stop improving.

\begin{table}[!t]
	\begin{center}
		\begin{tabular}{l|ccc}
			\textbf{LR}    			
			& \textbf{PSNR}$\uparrow$
			& \textbf{PSNR-$\mu$}$\uparrow$
			& \textbf{Epoch}\\ \hline			
			0.0001 	& 40.33 & 35.04 & 125\\ \hline
			0.0005-0.0001 	& \textbf{40.73} & \textbf{35.16} &  148 \\ 			
		\end{tabular}
	\end{center}
	\caption{\textbf{Quantitative Results} when training our final solution after 150 epochs, following the specifications of \cref{fig:lr} (Only first phase), against a model trained with a learning rate of 0.0001. Training with a slightly higher learning rate regime, the system is able to achieve higher accuracy faster and with room for a higher improvement. Results computed on our validation split.}
	\label{table:learning-rate}
\end{table}

\subsection{Low Resolution Branch Efficiency}
To quantify the efficiency gains by having a branch operating at a fourth of the resolution, we define an architecture (Ours *)  where both branches operate at full resolution. \Cref{table:efficiency-results} reports Runtime and GMACs (\#weights remain constant) against the final solution. The low resolution branch provides runtime efficiency gains of $\approx25\%$ and a GMACs reduction of $\approx30\%$.

\begin{table}[!t]
	\begin{center}
		\begin{tabular}{l|cc}
			\textbf{Method}    			
			& \textbf{Runtime (s)}$\downarrow$
			& \textbf{GMACs}$\downarrow$	\\ \hline
			
			Ours* &	0.972  & 2534.60 \\ \hline
			Ours  &	\textbf{0.738} & \textbf{1769.85} \\ 
		\end{tabular}
	\end{center}
	\caption{\textbf{Quantitative Results} when performing inference with an architecture where both branches operate at full resolution (Ours*) on images of 1060x1900 pixels.}
	\label{table:efficiency-results}
\end{table}

\subsection{Validation and Final results}
For completeness, in \Cref{table:results-phases} we include the validation phase results and the final test results of our model and the reference one. Our approach achieves better performance on both metrics, it has less trainable parameters and performs $\approx$40\% less operations. 

\begin{table}[!t]
	\begin{center}
		\begin{tabular}{l|cccc}
			\textbf{Method}    			
			& \textbf{PSNR}$\uparrow$
			& \textbf{PSNR-$\mu$}$\uparrow$
			& \textbf{\#weights}$\downarrow$
			& \textbf{GMACs}$\downarrow$
			\\ \hline			
			Baseline	& 38.34 & 36.86 & 1441283 & 2916.92\\ \hline
			Ours			 							& \textbf{38.50} & \textbf{36.91} & 1222035 & 1769.85 \\ \hhline{=|====}
			Baseline	& 37.60 & 37.02  & 1441283 & 2916.92\\ \hline
			Ours			 							& \textbf{38.49} & \textbf{37.11} & \textbf{1222035} & \textbf{1769.85} \\
			
		\end{tabular}
	\end{center}
	\caption{\textbf{Quantitative Results}. Validation split (top) and Test split (bottom) results against Baseline\cite{yan_attention_guided_2019} implementation from NTIRE 2022 HDR Challenge\cite{perezpellitero22}. }
	\label{table:results-phases}
\end{table}

\subsection{Qualitative Results}
We report qualitative results in \Cref{fig:qualitative-teaser} and \Cref{fig:qualitative-big} with samples from our validation split. From the different images, it can be seen that our method reproduces with high fidelity the ground truth samples. It also equalizes and even outperforms AHDR baseline. \cref{fig:qualitative-teaser} showcases a higher color fidelity while \cref{fig:qualitative-big-b} provides a much clear reconstruction when compared with the ground truth. 

Moreover, in \Cref{fig:qualitative-maps} we visualize the behavior of the main components of our system. In particular, we illustrate 3 input brackets $I_i, i = 1,2,3$, their respective activations from $[I_i, I^{h}_i], i = 1,3$ and the result of Deformable Convolutional Blocks and the Spatial Attention Blocks.  \texttt{feat0-h} is a single activation channel from $Z^{0}_3$, and \texttt{dconv-h} is the same channel after applying the Deformable Convlutional Block. On the other hand,  \texttt{feat1-l} is a single activation channel from $Z^{0}_1$ while \texttt{att-l} is the same channel after applying the Spatial Attention Block. \cref{fig:qualitative-maps} effectively illustrates how the Deformable Convolutional Block  suppresses details from the over-exposed regions of the table, while enhancing details from the background wall. For the low exposure bracket, \texttt{att-l} despite being at a fourth of the original resolution, is able to retain the well-exposed details from the table, while suppressing the over-exposed details from the background wall.

\section{Conclusion}
In this paper we have proposed DRHDR, an NTIRE 2022 HDR Challenge candidate solution for Track 2: Low-Complexity (Fidelity constrain). We unearth the benefit of exploiting spatially reduced feature representations for alleviating the high computational requirements of full resolution transformations. Despite the lower complexity, we demonstrate the ability of our system to provide accurate, ghost free HDR outputs with superior detail representation and higher efficiency when compared with baseline implementations.


\section*{Acknowledgments}
This work was supported by the Industrial Ph.D. program's financial support from Innovation Fund Denmark, through the project AIERE (Contract-No: 
9065-00099B).

{\small
\bibliographystyle{ieee_fullname}
\bibliography{egbib}
}

\end{document}